\documentclass{article}

\usepackage{microtype}
\usepackage[dvipsnames]{xcolor}
\usepackage{graphicx}
\usepackage{booktabs} 
\usepackage{subcaption}

\usepackage{hyperref}

\usepackage[accepted]{icml2024}

\usepackage{amsmath}
\usepackage{amssymb}
\usepackage{mathtools}
\usepackage{amsthm}

\usepackage[capitalize,noabbrev]{cleveref}

\theoremstyle{plain}

\theoremstyle{definition}

\theoremstyle{remark}

\usepackage[textsize=tiny]{todonotes}

\icmltitlerunning{HeteroLoRA}

\usepackage{xcolor, soul}
\usepackage{bbm}
\usepackage{algorithm}
\usepackage{algorithmic}
\usepackage{multirow}
\usepackage[UKenglish]{isodate}
\usepackage{pifont}
\usepackage{makecell}

\usepackage{pdfpages}
\usepackage[bottom]{footmisc}
\usepackage{natbib}
\usepackage{lipsum}

\begin{document}

\twocolumn[
\icmltitle{Unlocking the Global Synergies in Low-Rank Adapters}



\icmlsetsymbol{equal}{*}

\begin{icmlauthorlist}
\icmlauthor{Zixi Zhang}{cam}
\icmlauthor{Cheng Zhang}{ic}
\icmlauthor{Xitong Gao}{siat}
\icmlauthor{Robert D. Mullins}{cam}
\icmlauthor{George A. Constantinides}{ic}
\icmlauthor{Yiren Zhao}{ic}
\end{icmlauthorlist}

\icmlaffiliation{cam}{Department of Computer Science, University of Cambridge, Cambridge, United Kingdom}
\icmlaffiliation{ic}{Department of Electrical and Eletronic Engineering, Imperial College London, London, United Kingdom}
\icmlaffiliation{siat}{Shenzhen Institute of Advanced Technology, Chinese Academy of Sciences, Guangdong, China}

\icmlcorrespondingauthor{Zixi Zhang}{zz458@cam.ac.uk}
\icmlcorrespondingauthor{Cheng Zhang}{cheng.zhang122@imperial.ac.uk}
\icmlcorrespondingauthor{Xitong Gao}{xt.gao@siat.ac.cn}
\icmlcorrespondingauthor{Robert D. Mullins}{robert.mullins@cl.cam.ac.uk}
\icmlcorrespondingauthor{George A. Constantinides}{g.constantinides@imperial.ac.uk}
\icmlcorrespondingauthor{Yiren Zhao}{a.zhao@imperial.ac.uk}

\icmlkeywords{Machine Learning, ICML}

\vskip 0.3in
]



\printAffiliationsAndNotice{} 

\begin{abstract}


Low-rank Adaption (LoRA) has been the de-facto parameter-efficient fine-tuning technique for large language models. We present \textit{HeteroLoRA}, a light-weight search algorithm that leverages zero-cost proxies to allocate the limited LoRA trainable parameters across the model for better fine-tuned performance. In addition to the allocation for the standard LoRA-adapted models, we also demonstrate the efficacy of HeteroLoRA by performing the allocation in a more challenging search space that includes LoRA modules and LoRA-adapted shortcut connections. Experiments show that HeteroLoRA enables improvements in model
performance given the same parameter budge. For example, on MRPC, we see an improvement of 1.6\% in accuracy with similar training parameter budget. We will open-source our algorithm once the paper is accepted.

\end{abstract}

\section{Introduction}


\label{sec:motivation}

Recently, large language models (LLMs) have shown impressive performance in a range of natural language processing tasks~\cite{llmsurvey}. However, fine-tuning pre-trained language models (PLMs) is computationally and memory-intensive.
To mitigate this, \textit{parameter-efficient tuning} (PET) methods have been developed to fine-tune a small number of (extra) model parameters instead of the entire model~\citep{serialadapter}.

\begin{figure}
    \centering
    \includegraphics[width=\linewidth]{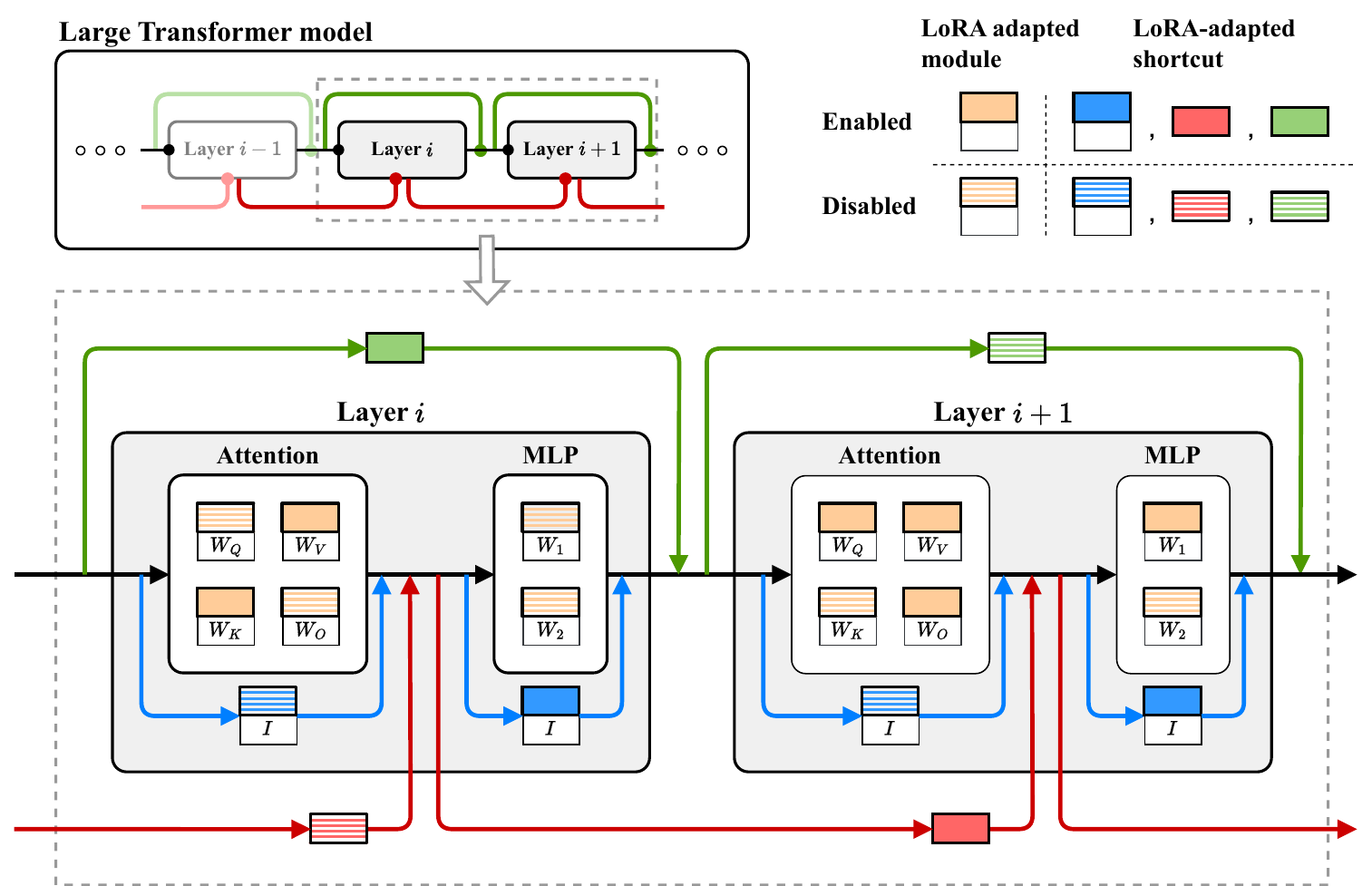}
    \caption{An illustration of the HeteroLoRA search space in a Transformer model. Given a fixed number of trainable parameters, HeteroLoRA finds an efficient heterogeneous LoRA configuration for a model on a specific task. Each of the standard LoRA module and LoRA-adapted shortcut can be enabled or disabled.}
    \label{fig: heterolora}
    \vspace{-15pt}
\end{figure}

Low-rank adaptation (LoRA) \citep{lora} is now the de-facto PET method. LoRA injects two low-rank matrices $A\in \mathbb{R}^{r\times d_\text{in}}$ and $B\in \mathbb{R}^{d_\text{out}\times r}$ with rank $r\ll\min(d_\text{in}, d_\text{out})$, to update the pre-trained weights $W \in \mathbb{R}^{d_\text{out}\times d_\text{in}}$.
Unlike full fine-tuning, LoRA updates only the injected $A$ and $B$ with the pre-trained weights $W$ unchanged. After fine-tuning, the update weights $\Delta W = BA$ fuse back to the pre-trained weights $W'=W+AB$, incurring no additional latency. LoRA achieves performance levels similar to full fine-tuning while drastically reducing memory usage.
We identify the following limitations of LoRA.
\begin{itemize}
    \item Existing methods configure LoRA modules within a model uniformly with the same rank $r$, thus each LoRA module consumes an identical number of trainable parameters, regardless of its potentially varying contributions to the overall model performance. 
    \item Current LoRA implementations predominantly adhere to the Transformer architecture. However, there has been limited exploration into extending the model architecture to enhance performance. This leads to the broader question of whether it is necessary to incorporate LoRA modules under these constraints and whether LoRA modules would be more effective with specific new connections, such as shortcut connections \cite{resnet, densenet}.
\end{itemize}

In this work, we introduce \textit{HeteroLoRA}, a new lightweight framework designed to autonomously allocate the LoRA module across the entire LLM given a parameter budget. Furthermore, we perform HeteroLoRA within an expanded search space including LoRA-adapted shortcut connections \citep{resnet} as illustrated in \Cref{fig: heterolora}. 

Specifically, we make the following contributions:


\begin{itemize}
    \item We propose HeteroLoRA, a novel LoRA configuration search algorithm to solve the rank allocation problem within a limited trainable parameter budget. HeteroLoRA leverages zero-cost proxies~\cite{zerocostproxy} to avoid the high cost of brute-force search. 
    \item We further prove the efficacy of the LoRA-adapted shortcut connection and combine it with HeteroLoRA to improve global synergies. The shortcuts suggested by HeteroLoRA enable more gains in model performance given the same parameter budget. For instance, on MRPC, we see an improvement of 1.6\% in accuracy with similar model size budgets.
\end{itemize}


\vspace{-1em}

\begin{table}[t]
\caption{Saliency Proxies for LoRA modules. We follow the definition of three zero-cost proxies, $s_\mathtt{snip}(\cdot)$ for SNIP, $s_\mathtt{synflow}(\cdot)$ for SYNFLOW, and $s_\mathtt{gradnorm}(\cdot)$ for GRAD-NORM, to build the saliency scores for LoRA modules ($S_\mathtt{snip}(\cdot)$, $S_\mathtt{synflow}(\cdot)$, and $S_\mathtt{gradnorm}(\cdot)$). A constant proxy is considered as random search baseline. Detailed introduction to zero-cost proxies~\cite{zerocostproxy} is included in~\Cref{app:proxies}.}
\label{tab:proxies}
\vspace{1em}
\begin{small}
\resizebox{\linewidth}{!}{
    \begin{tabular}{@{}ll@{}}
    \toprule
    Proxy      & Saliency score of LoRA-adapted module                        \\ \midrule
    Constant   & $S_\mathtt{constant}(M) = 1$     \\[2ex]
    GRAD-NORM~\cite{zerocostproxy} & $S_\mathtt{gradnorm}(M) = s_\mathtt{gradnorm}(A) + s_\mathtt{gradnorm}(B)$  \\[2ex]    
    SNIP~\cite{snip}       & $S_\mathtt{snip}(M)=\sum\limits_{\theta\in A} s_\mathtt{snip}(\theta) + \sum\limits_{\phi\in B} s_\mathtt{snip}(\phi)$     \\[2ex]
    SYNFLOW~\cite{synflow}    & $S_\mathtt{synflow}(M) = \sum\limits_{\theta\in A} s_\mathtt{synflow}(\theta) + \sum\limits_{\phi\in B} s_\mathtt{synflow}(\phi)$       \\
    \bottomrule
    \vspace{-2em}
    \end{tabular}
}
\end{small}
\end{table}
\section{HeteroLoRA}

We adopt zero-cost proxies to estimate the importance of LoRA modules in~\Cref{sec: lora search}, and discuss two ways to integrate the HeteroLoRA search into the existing PET pipeline in~\Cref{sec:method:static-or-dynamic}. In~\Cref{sec:method:add-shortcuts}, we introduce LoRA-adapted shortcuts to enable exploration of global synergies.

\subsection{Saliency Estimation using Zero-Cost Proxies}
\label{sec: lora search}


Given a limited number of active LoRA modules, turning on a subset of modules could potentially be more effective. For instance, turning on all LoRA modules with $r=2$ vs.\ turning on $25\%$ of all modules with $r=8$, the latter may achieve higher model performance. We consider such LoRA configuration assignment as a ``LoRA rank allocation'' problem and propose HeteroLoRA to solve it.




We estimate the saliency (importance) of a single LoRA module as a reference for HeteroLoRA searches. Modules with higher saliency scores will be enabled during training, and tie-breaking will be done by uniform random sampling. Three saliency proxies plus a random allocation baseline are shortlisted in~\Cref{tab:proxies}. The detailed introduction to SNIP~\cite{snip}, SYNFLOW~\cite{synflow}, and GRAD-NORM~\cite{zerocostproxy} are included in the \Cref{app:proxies}.

Note that the saliency proxies are applied to the whole LoRA module $W'$ instead of $\Delta W = BA$ for two reasons. First, at the start of training, the update component $\Delta W = BA$ is initialised as zeros, hence saliencies do not make sense by then. Second, the update component $\Delta W$ has a strong correlation with the pre-trained weight $W$, indicating that the features that $\Delta W$ amplified are already in $W$ \citep{lora}. Therefore, it is reasonable to include the pre-trained weight in the saliency measurements for deciding the ``on/off'' of the LoRA modules.

\begin{figure}[t]
    \centering
    \begin{subfigure}{\linewidth}
     \includegraphics[width=\textwidth]{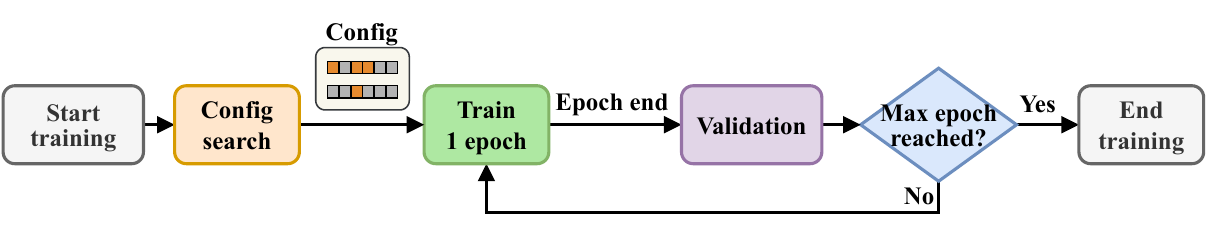}
     \caption{Static HeteroLoRA} \label{fig: static heterolora pipeline}
    \end{subfigure}%
    \\
    \centering
    \begin{subfigure}{\linewidth}
     \includegraphics[width=\textwidth]{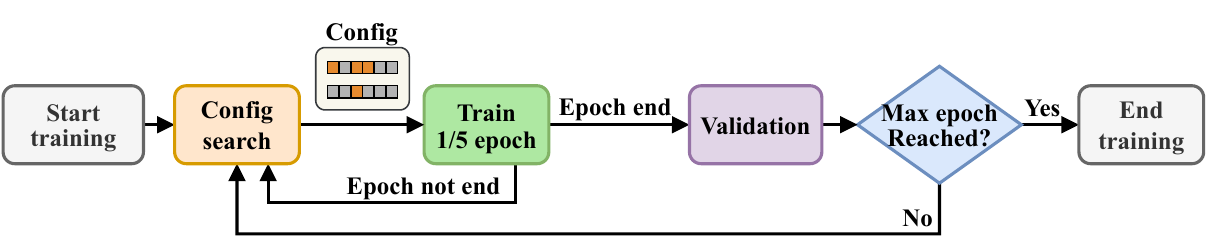}
     \caption{Dynamic HeteroLoRA} \label{fig: dynamic heterolora pipeline}
    \end{subfigure}
    \caption{Training pipeline of (a) static HeteroLoRA, where LoRA modules are enabled/disabled at the start of training, and (b) dynamic HeteroLoRA, where LoRA modules are enabled/disabled periodically, \textit{e.g.}, every $1/5$ epoch, during the training.} 
    \vspace{-1em}
    \label{fig: heterolora pipelines}
\end{figure}

\subsection{Static or Dynamic?}
\label{sec:method:static-or-dynamic}

\paragraph{Static HeteroLoRA} A straightforward way to incorporate HeteroLoRA search into the training pipeline is to compute the saliency proxy at the beginning of training, enabling or disabling LoRA modules accordingly.
This is usually applied on a handful of training samples at the start that only introduce minimal search cost, taking around 10\% of one epoch training.
We then maintain the same rank allocation throughout training, as depicted in \Cref{fig: static heterolora pipeline}. This approach mirrors zero-cost NAS~\cite{zerocostproxy}, where lightweight search for optimal configurations are conducted initially, followed by complete fine-tuning. 



\paragraph{Dynamic HeteroLoRA} After several training steps, the optimizer may find that some enabled LoRA modules are not as important as measured initially. Therefore, we introduce dynamic HeteroLoRA, which periodically updates the rank allocation at the start of each training epoch, as shown in \Cref{fig: dynamic heterolora pipeline}. Dynamic HeteroLoRA offers an opportunity to inspect the importance of each LoRA module through the frequency it has been enabled.


\subsection{Extending the Search Space with LoRA-Adapted Shortcut Connections}
\label{sec:method:add-shortcuts}



We introduce LoRA-adapted shortcut connections to extend the search space, which is later integrated with HeteroLoRA to foster global synergies between LoRA modules. A LoRA-style low-rank linear transformation is applied to each shortcut:
$$W = W_0 + \frac{\alpha}{r}BA$$
where $W_0$ is the initial weight of the linear projection, depending on the type of the shortcut; $\alpha$ is a pre-defined scaling factor; and $r$ is the rank of $A$ and $B$. $A$ and $B$ are initialised similarly to LoRA modules to ensure $W = W_0$ at the start of training. We refer to a combination of $\langle W_0, A, B \rangle$ as a ``shortcut module''.
A layer normalisation \citep{layernorm} is appended after the addition of the shortcut to improve the training stability.

\begin{figure}[t]
    \centering
    \includegraphics[width=0.8\linewidth]{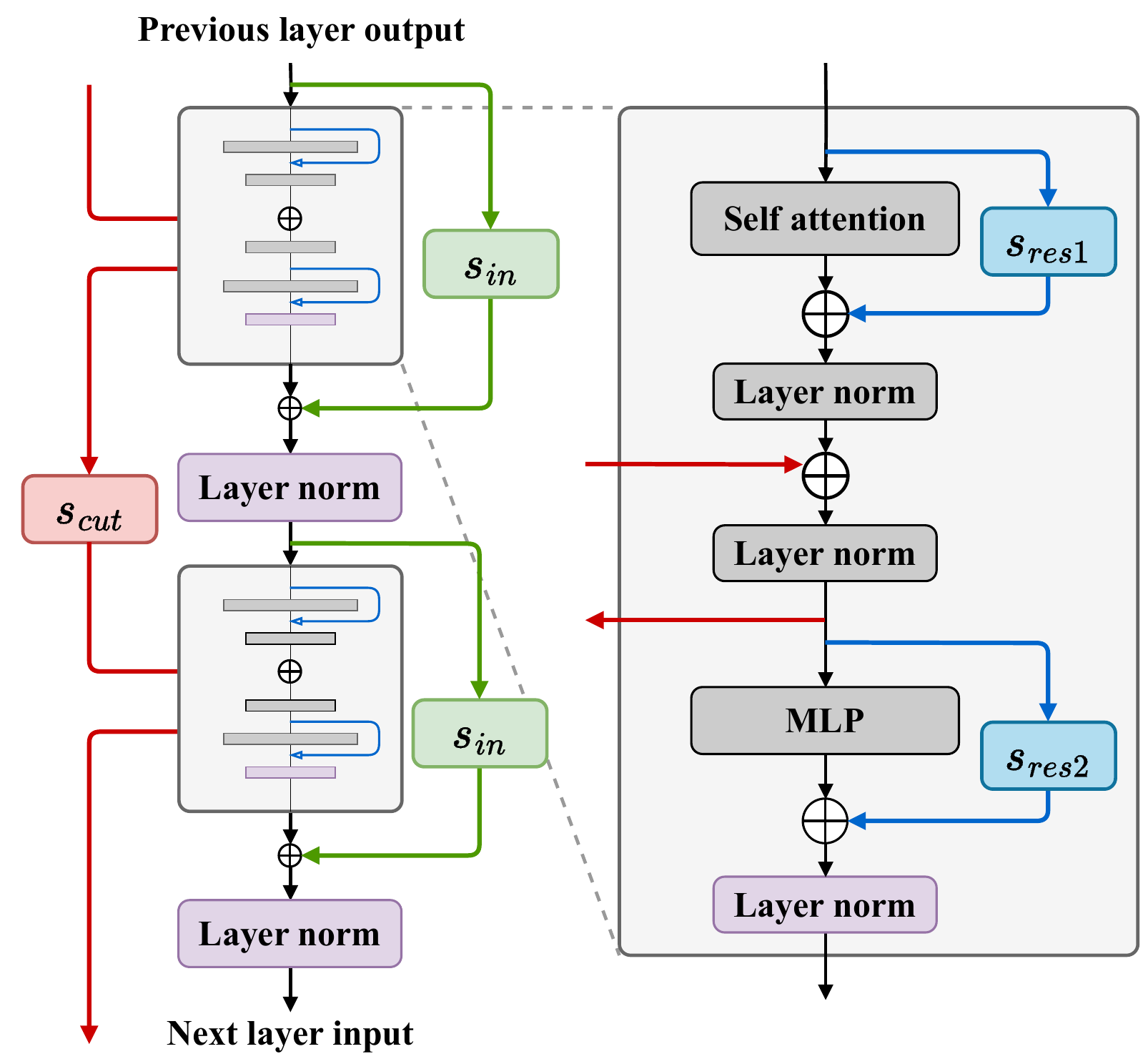}
    \caption{LoRA-adapted shortcut architecture on two Transformer layers with post-layer-normalisation. \textcolor{RoyalBlue}{Blue} blocks are the residual shortcuts $s_\text{res1}$ and $s_\text{res2}$. \textcolor{Green}{Green} blocks are the $s_\text{in}$ same-layer style ``cross-layer'' shortcuts. \textcolor{BrickRed}{Red} blocks are the $s_\text{cut}$ ``cut-layer'' style cross-layer shortcuts.}
    \vspace{-1em}
    \label{fig: shortcut arch}
\end{figure}

As shown in \Cref{fig: shortcut arch}, we focus on two types of shortcut connections, residual shortcut and cross-layer shortcut, to keep the search space tractable: 
\vspace{-0.5em}
\begin{enumerate}
    \item Residual shortcut: Shortcut is applied to the micro-architecture by replacing the two original residual connections within the Transformer block (as the \textcolor{RoyalBlue}{blue} blocks in \Cref{fig: shortcut arch}). We refer to them as ``residual shortcuts'' $s_\text{res1}$ and $s_\text{res2}$. The initial weight $W_0$ of these shortcut modules is the identity matrix $I$. 
    \item Cross-layer shortcut: Shortcut connection is applied to the macro-architecture by linking two points at different Transformer blocks. We refer to this as the ``cross-layer shortcut''. A cross-layer shortcut skips multiple Transformer blocks. We concentrate on cross-layer shortcuts that skip one Transformer block for simplicity, as the \textcolor{Green}{green} blocks ($s_{\text{in}}$) and the \textcolor{BrickRed}{red} blocks ($s_{\text{cut}}$) in \Cref{fig: shortcut arch}. 
The initial weights $W_0$ of the cross-layer shortcut modules are initialised to zeros because these shortcuts do not exist in the original architecture. 
\end{enumerate}

\vspace{-2em}
\section{Experiments}

We outline our basic experiment setup in \Cref{sec:exp:setup}. In \Cref{sec: saliency heuristic cmp}, we identify the most promising saliency proxy and verify that HeteroLoRA achieves a better performance than the standard homogeneous LoRA. In \Cref{sec: shortcut arch r-search result}, we demonstrate that LoRA-adapted shortcuts enable an additional performance gain. In~\Cref{sec: dyrealloc result} we use HeteroLoRA to allocate the rank in a search space that includes both standard LoRA moduels and LoRA-adapted shortcuts, highlighting the effecacy of HeteroLoRA.

\vspace{-0.5em}
\subsection{Experiment Setup}
\label{sec:exp:setup}

We perform the experiments on OPT-350M~\cite{opt} over GLUE subsets MRPC, RTE, and SST-2~\cite{glue}. For each experiment, the median and standard deviation of the performance are calculated over five independent runs of different random seeds. We apply LoRA to the query layer $W^Q$ and the value layer $W^V$ as experiments show that applying these LoRA modules to these two layers effectively improves model performance (see \Cref{app:primary-lora-experiments}). We use a fixed number of trainable parameters for all groups in the same experiment subsection. Detailed rank and training hyperparameters are summarised in \Cref{appendix:hyperparam}.

\vspace{-0.5em}
\subsection{Determining Proxy and Training Strategy of HeteroLoRA}
\label{sec: saliency heuristic cmp}

\Cref{tab:lora_alloc_mrpc} shows the comparison of the four proxies under static and dynamic HeteroLoRA. In each experiment, $25\%$ of LoRA modules are enabled with $r=8$. The eight combinations are also compared to a baseline, in which all LoRA modules are enabled with $r=2$, so the numbers of trainable parameters are the same. 
We observe that \textit{dynamic HeterLoRA achieves better performances than static HeteroLoRA}, with {GRAD-NORM} performing the best and surpassing the baseline. Therefore, we use dynamic HeteroLoRA with the {GRAD-NORM} proxy in the following experiments.


\begin{table}
    \caption{Performance of the salience proxies with static and dynamic HeteroLoRA. Accuracy on MRPC and the difference with the baseline performance are reported. The baseline, in which all LoRA modules are enabled with rank $r=2$, achieves $83.8\%$ accuracy. We observe that the combination of \texttt{GRAD-NORM} and dynamic HeteroLoRA achieves the highest accuracy.}
    \label{tab:lora_alloc_mrpc}
    \vspace{1em}
    \centering
    \resizebox{\linewidth}{!}{%
    \begin{small}
        \begin{tabular}{lcccc}
        \toprule
        $r=8$          & {CONSTANT} & {GRAD-NORM}   & {SNIP} & {SYNFLOW} \\ \midrule
        Static & 83.8 (+0.0)       & 82.8 (-1.0)          & 82.8 (-1.0)   & 78.7 (-5.1)      \\
        Dynamic & 82.4 (-1.4)       & \textbf{84.1 (+0.3)} & 82.4 (-1.4)   & 82.4 (-1.4)      \\ \bottomrule
        \end{tabular}
    \end{small}
    }
    
\end{table}

\begin{table}[!t]
    \caption{Performance gain of LoRA-adapted shortcut connections. L-only is the LoRA-only baseline. $r_S=r_L$ denotes the model in which both LoRA modules and LoRA-adapted shortcuts are applied with the same rank. $r_S$$>$$r_L$ represent the model in which the LoRA module rank is fixed and the rest of ranks are allocated to the shortcut. We observe that the LoRA-adapted shorcuts combined with LoRA modules achieve higher accuracy than the LoRA-only group given the same number of trainable parameters. 
    }
    \label{tab:r-search}
    \vspace{1em}
    \centering
    \resizebox{\linewidth}{!}{%
    \begin{small}
\begin{tabular}{@{}clcccc@{}}
\toprule
        \#Trainable            & Group                         & MRPC                & RTE                 & SST-2                                    & Avg.          \\ \midrule
        \multirow{3}{*}{2.3M} & \multicolumn{1}{l|}{L-only}     & 83.4 ± 1.1          & 72.1 ± 1.2          & \multicolumn{1}{c|}{93.4 ± 0.4}          & 92.1          \\
         & \multicolumn{1}{l|}{$r_S=r_L$} & \textbf{84.6 ± 0.5} & \textbf{73.5 ± 1.4} & \multicolumn{1}{c|}{\textbf{93.9 ± 0.3}} & \textbf{92.6} \\
                              & \multicolumn{1}{l|}{$r_S > r_L$}     & 84.6 ± 1.9          & 70.9 ± 2.2          & \multicolumn{1}{c|}{93.7 ± 0.3}          & 92.4          \\ \midrule
        \multirow{3}{*}{9.4M} & \multicolumn{1}{l|}{L-only}     & 83.4 ± 0.4          & 69.4 ± 3.2          & \multicolumn{1}{c|}{93.3 ± 0.5}          & 91.9          \\
                              & \multicolumn{1}{l|}{$r_S=r_L$} & 83.8 ± 0.6          & \textbf{72.6 ± 1.3} & \multicolumn{1}{c|}{\textbf{94.1 ± 0.7}} & \textbf{92.7} \\
                              & \multicolumn{1}{l|}{$r_S > r_L$}     & \textbf{84.1 ± 0.9} & 71.8 ± 3.0          & 93.8 ± 0.3                               & 92.5          \\ \bottomrule
        \end{tabular}
    \end{small}
    }
    \vspace{-1em}
\end{table}

\subsection{Verifying Performance Gain of Shortcuts}
\label{sec: shortcut arch r-search result}
\vspace{-0.5em}


We conduct controlled experiments in various configurations to validate the effectiveness of shortcut connections. \Cref{tab:r-search} presents two shortcut setups under 2.3M and 9.4M trainable parameters:

\vspace{-1em}
\begin{itemize}
    \item L-only: Only the standard LoRA is applied to the model. This group serves as the baseline.
    \item $r_S = r_L$: Both standard LoRA and LoRA-style shortcuts are applied to the model and the rank of shortcuts is the same as the standard LoRA.
    \item $r_S > r_L$: Both LoRA and shortcut are applied to the model, but the LoRA module rank is fixed and the rest of ranks are allocated to the shortcut.
\end{itemize}

\vspace{-0.5em}
We ensure that the three groups have the same number of trainable parameters for a fair comparison. The detailed experiment setup is included in \Cref{appendix:hyperparam}.
We observe that the shortcut-adapted architecture generally outperforms the LoRA-only architecture, meanwhile with a more prominent advantage as the budget becomes larger. This observation indicates that the linear projections on the shortcuts have larger ``intrinsic ranks'' than the LoRA update matrices. When performance has ``saturated'' in LoRA modules, shortcuts foster further performance improvement by developing global synergies across layers.

\subsection{Dynamic HeteroLoRA with Extended Search Space}
\label{sec: dyrealloc result}

Finally, we integrate the components discussed above.
We combine Dynamic HeteroLoRA and {GRAD-NORM}, as explored in \Cref{sec: saliency heuristic cmp}, thereby embracing the expanded search space as inferred from the findings in \Cref{sec: shortcut arch r-search result}.

\begin{table}
    \caption{Dynamic HeteroLoRA combined with LoRA-adapted shortcuts. The S \& L baselines have all modules enabled. The DH \& S \& L have $25\%$ of LoRA models and LoRA-adapted shortcuts enabled. They have the same number of trainable parameters for a fair comparison. We observe that DH \& S \& L outperforms the baseline, meaning HeteroLoRA finds a more optimal rank allocation than the homogeneous baseline in a challenging search space including both LoRA modules and shortcuts.}
    \label{tab:dyrealloc}
    \vspace{1em}
    \centering
    \resizebox{\linewidth}{!}{%
        \begin{small}
        \begin{tabular}{@{}clcccc@{}}
        \toprule
        Rank budget & Setup      & MRPC                & RTE               & SST-2             & Avg. \\ \midrule 
        \multirow{2}{*}{\makecell{r=2 for S \& L}} & S \& L      & 83.7 ± 0.8          & \textbf{73.4 ± 2.2} & 93.6 ± 0.5      &  83.5   \\
                                   & DH \& S \& L & \textbf{84.3 ± 1.0} & 72.9 ± 1.8          & \textbf{93.9 ± 0.5}  & \textbf{83.7} \\ \midrule
        \multirow{2}{*}{\makecell{r=8 for S \& L}} & S \& L     & 84.6 ± 0.5          & 73.5 ± 1.4          & \textbf{93.8 ± 0.4} &  84.0 \\
                                   & DH \& S \& L & \textbf{85.0 ± 1.6} & \textbf{73.6 ± 1.1} & 93.6 ± 0.1       &  \textbf{84.1} \\ \bottomrule
        \end{tabular}
        \end{small}
    }
\end{table}


\Cref{tab:dyrealloc} compares the LoRA-adapted model with/without Dynamic LoRA under the same trainable parameter budget:
\vspace{-0.5em}
\begin{itemize}
    \item S \& L denotes all the LoRA modules and LoRA-adapted shortcuts are enabled with the same rank.
    \item For group DH \& S \& L, HeteroLoRA sorts the saliency scores of standard LoRA and LoRA-adapted shortcuts to determine which module to enable/disable.
\end{itemize}

\begin{figure}[t]
    \centering
    \includegraphics[width=\linewidth]{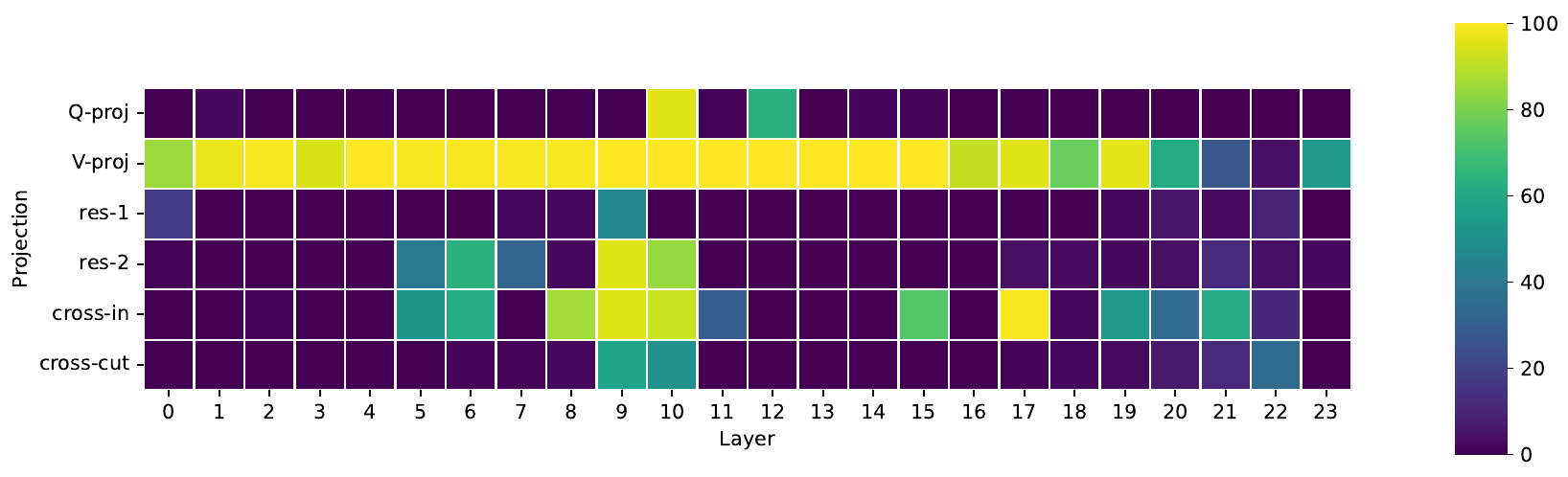}
    \caption{Frequency of linear projections in every model layer being enabled in the dynamic HeteroLoRA trained on MRPC for LoRA and shortcut modules. A noticeable preference for value projections over query projections indicates that the value update generally contributes more to the fine-tuned performance. }
    \label{fig:dyrealloc_heatmap_mrpc}
\end{figure}

As illustrated in \Cref{tab:dyrealloc}, we observe that dynamic HeteroLoRA further improves model performance over S \& L, indicating the HeteroLoRA finds a more optimal rank allocation.
\Cref{fig:dyrealloc_heatmap_mrpc} displays the frequency of each LoRA or shortcut module being enabled over the 20 training epochs on MRPC. The frequency of each LoRA module denotes its importance to performance; the frequency of each shortcut module characterises its efficacy to global synergies. A noticeable preference for value projections over query projections indicates that the value transformation updates generally contribute more to the performance. Move results are available in \Cref{app:additional results}. 


\section{Conclusions}

We propose dynamic HeteroLoRA, a framework automatically determining the ``on/off'' for the LoRA modules in LLM fine-tuning. Then we verify that LoRA-adapted shortcuts improve model performance. In the end, we demonstrate that dynamic solvesoLoRA effectively solve the rank allocation problem in a challenging search space including both LoRA modules and shortcuts. HeteroLoRA offers a cost-effective way to allocate trainable parameters within a limited training budget.

\bibliography{sample}

\newpage
\appendix
\onecolumn
\section{Primary LoRA Experiments}
\label{app:primary-lora-experiments}

To make the search space manageable, we first decide which linear layers in attention are helpful if they are LoRA-adapted. We sweep the LoRA combinations and fine-tune them on MRPC. As shown in~\Cref{tab:lora_abla_proj1,,tab:lora_abla_proj2}, we find that applying LoRA to $W_Q$ and $W_V$ is most helpful for improving model performance.

\begin{table}[h]
    \caption{Performance of LoRA applied to different combinations of linear projections in the attention submodule under the same numbers of trainable parameters. Adapting $W^V$ and $(W^Q, W^V)$ perform the best.}
    \label{tab:lora_abla_proj1}
    \centering
    \begin{tabular}{c|ccccccc}
    \toprule
    Projections & $W^Q$ & $W^K$ & $W^V$                           & $W^O$                & $W^Q, W^K$ & $W^Q, W^V$ & \begin{tabular}[c]{@{}c@{}} $W^Q, W^K,$ \\ $W^V, W^O$ \end{tabular} \\
    Rank $r$ & 8 & 8 & 8 & 8 & 4 & 4 & 2  \\
    \midrule
    MRPC (acc) & $80.2{\pm 1.6}$  & $80.3{\pm 0.5}$ & $\mathbf{84.3{\pm 1.3}}$ & $83.2{\pm 0.1} $ & $81.6{\pm 1.1}$ & $\mathbf{84.1{\pm 1.1}}$ & $83.3{\pm 1.8}$ \\    
    \bottomrule
    \end{tabular}
\end{table}

\begin{table}[h]
    \caption{Comparison of LoRA applied to attention linear projections, FFN linear projections, and both. Due to the larger feed-forward network module dimension, adapting FFN projections $W_1$ or $W_2$ costs more trainable parameters than the attention projections.}
    \label{tab:lora_abla_proj2}
    \centering
    \begin{tabular}{>{\centering\arraybackslash}m{0.16\linewidth}|>{\centering\arraybackslash}m{0.14\linewidth}|>{\centering\arraybackslash}m{0.1\linewidth}>{\centering\arraybackslash}m{0.1\linewidth}>{\centering\arraybackslash}m{0.14\linewidth}|>{\centering\arraybackslash}m{0.2\linewidth}}
    \toprule
    Projections & $W^Q, W^V$ & $W_1$ & $W_2$ & $W_1, W_2$ & $W^Q, W^V, W_1, W_2$ \\
    Rank $r$       & 4  & 8     & 8     & 4          & 2 \\
    \# Trainable   & 394K & \multicolumn{3}{c|}{984K}                       & 689K \\
    \midrule
    MRPC (acc) & $\mathbf{84.1{\pm 1.1}}$ & $83.3{\pm 0.6}$ & $82.3{\pm 1.2}$ & $84.0{\pm 0.9}$ & $83.5{\pm 0.5}$ \\
    \bottomrule
    \end{tabular}
\end{table}

\section{Hyperparameters}
\label{appendix:hyperparam}

For experiments with different model configurations and different datasets, the optimal training hyperparameters may vary. Due to the expensive computational cost, searching for the optimal hyperparameters in every experiment is infeasible. Therefore, learning rate searches are conducted for experiments while other hyperparameters are set by referencing the original RoBERTa paper \citep{roberta} and LoRA paper \citep{lora}. 

For experiments on the LoRA-searching training pipelines, the optimal hyperparameters tuned independently by beam search on \textit{MRPC} are applied to all datasets, except for the learning rate, which is searched on each dataset respectively. 

\subsection{LoRA modules combined with LoRA-Adapted Shortcuts}
\label{app:hyperparam:shortcut-and-lora}

In the dynamic HeteroLoRA training, a search is conducted every $1/5$ training epoch, in which the \texttt{GRAD-NORM} saliency score is evaluated over 32 batches of training data for each LoRA or shortcut module, and the modules ranking top $25\%$ are enabled with rank $r=8$ until the next HeteroLoRA search.

As shortcut connections have shown their effectiveness at larger ranks in \Cref{sec: shortcut arch r-search result}, we further evaluate dynamic HeteroLoRA with LoRA and shortcut modules enabled with rank $r=32$. The two HeteroLoRA training setups are compared to two baselines with all modules enabled but with $1/4$ ranks respectively, so the numbers of trainable parameters at any time in the training are equivalent respectively.

\begin{table*}[h]
    \caption{The hyperparameters for training on datasets from the GLUE benchmark.}
    \label{tab: hyperparam core}
    \centering
    \begin{small}
    \begin{tabular}{cc|ccccc}
    \toprule
    Method                            & Dataset       & MNLI & MRPC & QQP  & RTE  & SST-2 \\
    \midrule
     & Optimiser & \multicolumn{5}{c}{AdamW} \\
    \midrule
    \multirow{5}{*}{OPT-350M FT}      & Batch size    & 32   & 16   & 32   & 16   & 32    \\
                                      & \# Epochs     & 5    & 10   & 5    & 10   & 5     \\
                                      & Learning rate & 1e-5 & 2e-5 & 1e-5 & 1e-5 & 1e-5  \\
                                      & Weight decay  & \multicolumn{5}{c}{0.01}          \\
                                      & Max Seq. Len. & \multicolumn{5}{c}{512}           \\
    \midrule
    \multirow{5}{*}{RoBERTa-large FT} & Batch size    & 32   & 16   & 32   & 16   & 32    \\
                                      & \# Epochs     & 5    & 10   & 5    & 10   & 5     \\
                                      & Learning rate & 2e-5 & 1e-5 & 1e-5 & 1e-5 & 1e-5  \\
                                      & Weight decay  & \multicolumn{5}{c}{0.01}          \\
                                      & Max Seq. Len. & \multicolumn{5}{c}{512}           \\
    \midrule
    \multirow{7}{*}{OPT-350M LoRA}      & Batch size    & \multicolumn{5}{c}{8}               \\
                                    & \# Epochs     & 10    & 20    & 10   & 20   & 10    \\
                                    & Learning rate & 1e-4  & 2e-4  & 2e-4 & 3e-4 & 1e-4  \\
                                    & Weight decay  & \multicolumn{5}{c}{0.01}            \\
                                    & Max Seq. Len. & \multicolumn{5}{c}{512}             \\
                                    & LoRA config.  & \multicolumn{5}{c}{$r_Q = r_V = 8$} \\
                                    & LoRA $\alpha$ & \multicolumn{5}{c}{16}              \\
    \midrule
    \multirow{7}{*}{RoBERTa-large LoRA} & Batch size    & \multicolumn{5}{c}{8}               \\
                                    & \# Epochs     & 10    & 20    & 10   & 20   & 10    \\
                                    & Learning rate & 1e-4  & 2e-4  & 1e-4 & 4e-4 & 1e-4  \\
                                    & Weight decay  & \multicolumn{5}{c}{0.01}            \\
                                    & Max Seq. Len. & \multicolumn{5}{c}{512}             \\
                                    & LoRA config.  & \multicolumn{5}{c}{$r_Q = r_V = 8$} \\
                                    & LoRA $\alpha$ & \multicolumn{5}{c}{16}              \\
    \midrule
    \multirow{7}{*}{Gemma-2B LoRA}  & Batch size    & \multicolumn{5}{c}{8}               \\
                                    & \# Epochs     & 5    & 20    & 5   & 20   & 10      \\
                                    & Learning rate & 5e-5  & 2e-4  & 1e-4 & 5e-4 & 1e-4  \\
                                    & Weight decay  & \multicolumn{5}{c}{0.01}            \\
                                    & Max Seq. Len. & \multicolumn{5}{c}{512}             \\
                                    & LoRA config.  & \multicolumn{5}{c}{$r_Q = r_V = 8$} \\
                                    & LoRA $\alpha$ & \multicolumn{5}{c}{16}              \\
    \bottomrule
    \end{tabular}
    \end{small}
\end{table*}
\begin{table*}[h]
    \caption{The hyperparameters for training on Alpaca and datasets from the SuperGLUE benchmark.}
    \label{tab: hyperparam extension}
    \centering
    \begin{small}
    \begin{tabular}{cc|cccc}
    \toprule
    Method                         & Dataset       & Alpaca   & BoolQ   & CB     & COPA  \\
    \midrule
                                   & Optimiser     & \multicolumn{4}{c}{AdamW}           \\
    \midrule
    \multirow{5}{*}{OPT-350M FT}   & Batch size    & 32       & 32      & 16     & 16    \\
                                   & \# Epochs     & 10       & 10      & 10     & 10    \\
                                   & Learning rate & 1e-5     & 2e-5    & 5e-5   & 2e-5  \\
                                   & Weight decay  & \multicolumn{4}{c}{0.01}            \\
                                   & Max Seq. Len. & \multicolumn{4}{c}{512}             \\
    \midrule
    \multirow{7}{*}{OPT-350M LoRA} & Batch size    & \multicolumn{4}{c}{8}               \\
                                   & \# Epochs     & 20       & 10      & 20     & 20    \\
                                   & Learning rate & 1e-4     & 2e-4    & 1e-3   & 1e-3  \\
                                   & Weight decay  & \multicolumn{4}{c}{0.01}            \\
                                   & Max Seq. Len. & \multicolumn{4}{c}{512}             \\
                                   & LoRA config.  & \multicolumn{4}{c}{$r_Q = r_V = 8$} \\
                                   & LoRA $\alpha$ & \multicolumn{4}{c}{16}              \\
    \bottomrule
    \end{tabular}
    \end{small}
\end{table*}
\begin{table*}[h]
    \caption{The hyperparameters for experiments on shortcut-adapted models.}
    \label{tab: hyperparam shortcut}
    \centering
    \begin{tabular}{cc|>{\centering\arraybackslash}m{0.12\linewidth}>{\centering\arraybackslash}m{0.12\linewidth}>{\centering\arraybackslash}m{0.12\linewidth}}
    \toprule
    Method & Dataset           & MRPC       & RTE       & SST-2      \\
    \midrule
           & Optimiser         & \multicolumn{3}{c}{AdamW}           \\
    \midrule
    \multirow{9}{*}{\begin{tabular}[c]{@{}c@{}}OPT-350M\\ with shortcuts\end{tabular}} &
      Batch size &
      \multicolumn{3}{c}{8} \\
           & \# Epochs         & 20         & 20        & 10         \\
           & Learning rate     & 2e-4       & 2e-4      & 1e-4       \\
           & Weight decay      & \multicolumn{3}{c}{0.01}            \\
           & Max Seq. Len.     & \multicolumn{3}{c}{512}             \\
           & LoRA config.      & \multicolumn{3}{c}{$r_Q = r_V = 8$} \\
           & LoRA $\alpha$     & \multicolumn{3}{c}{16}              \\
     &
      Shortcut config. &
      \multicolumn{3}{c}{$r_\text{res1} = r_\text{res2} = r_\text{in} = r_\text{cut} = 8$} \\
           & Shortcut $\alpha$ & \multicolumn{3}{c}{4}               \\
    \bottomrule
    \end{tabular}
\end{table*}

\subsection{Training hyperparameters}
\label{app:hyperparam}

Experiments were run on five random seeds (0, 13, 42, 87, 100). The medians on every performance metric were reported. All other experiments were run on three random seeds (0, 13, 42), and the averaged results were reported.
\Cref{tab: hyperparam core} and \Cref{tab: hyperparam extension} display the training hyperparameters. The learning rates were determined through hyperparameter searches from 1e-5 to 5e-3; other hyperparameters were decided according to previous works \citep{roberta} and \citep{lora}.

\subsection{Saliency hyperparameters}
For comparison between saliency proxies, the following hyperparameters of the proxies were used:
\begin{itemize}
    \item \texttt{CONSTANT}: no hyperparameter required.
    \item \texttt{SNIP}: the score was evaluated on the first 32 batches from the training set.
    \item \texttt{SYNFLOW}: no hyperparameter required.
    \item \texttt{GRAD-NORM}: the score was evaluated on the first 32 batches from the training set.
\end{itemize}

Experiments were conducted to compare different choices of the number of training batches to use for \texttt{SNIP} and \texttt{GRAD-NORM}. On MRPC, using 8 batches, 32 batches, and the entire training set produced the same training curve. Following the previous work \citep{zerocostproxy}, I used 32 training batches for the later experiments.

Furthermore, two hyperparameters were involved in the HeteroLoRA strategies:
\begin{itemize}
    \item Enable rate: the percentage of LoRA and shortcut modules enabled at any time in the training was set to 25\%.
    \item Frequency of dynamic HeteroLoRA search: for all experiments in \Cref{sec: saliency heuristic cmp} and \Cref{sec: dyrealloc result}, HeteroLoRA search was conducted 5 times per training epoch. Further ablation experiment results are shown in \Cref{app:additional results}.
\end{itemize}

The training hyperparameters in \Cref{sec: saliency heuristic cmp} followed \Cref{tab: hyperparam core} except the learning rates, which were searched across 5e-5 to 1e-3 respectively for each proxy and training strategy.

The learning rates for the shortcut-adapted models were searched from 5e-5 to 5e-4. Moreover, the scaling factor $\alpha$ of shortcut modules was searched across 1 to 16. The best hyperparameters and configurations, as in \Cref{tab: hyperparam shortcut}, were used for the two series of shortcut-adapted models in \Cref{sec: shortcut arch r-search result} and the dynamic HeteroLoRA training in \Cref{sec: dyrealloc result} unless particularly specified.

\section{Zero-Cost Proxies}
\label{app:proxies}

The detailed definition of zero-cost proxies for the LoRA-adapted module/shortcut and trainable parameters are defined as follows

\subsection{CONSTANT}\label{app:proxy:constant} 
A baseline proxy is designed as assigning score $S_\mathtt{constant}(M) = 1$ to every LoRA module $M$. This enforces tie-breaking on all LoRA modules, so uniform random sampling is performed in every HeteroLoRA search.

\subsection{SNIP}\label{app:proxy:snip} 
The SNIP \citep{snip} proxy aims to find the elements that degrade the performance the least when removed. It uses a weight mask $C\in \{0,1\}^m$ applied to each block of parameters, with $0$ at the positions of disabled parameters and $1$ at the position of active parameters, and computes the loss gradient to the mask variables over a few minibatches of training data $\mathcal{D}$:
$$s_\mathtt{snip}(\theta) = \frac{\partial \mathcal{L}(\mathcal{D}; C\odot W)}{\partial c_\theta}$$
where $c_\theta$ denotes the weight mask variable corresponding to parameter $\theta$. In HeteroLoRA, since the LoRA rank allocation regards each LoRA module as a unit, we fill the weight mask $C$ for each LoRA module with ones, and extend the saliency of a single parameter $s(\theta)$ the saliency of a LoRA module $M$ by summation:
\begin{align*}
    S_\mathtt{snip}(M) &= \sum_{\theta\in M} s_\mathtt{snip}(\theta) \\
    &= \sum_{\theta\in A} s_\mathtt{snip}(\theta) + \sum_{\phi\in B} s_\mathtt{snip}(\phi)
\end{align*}

\subsection{SYNFLOW}\label{app:proxy:synflow} 
A minibatch of inputs of ones is fed to the model with weights taken as their absolute values. The \texttt{SYNFLOW} \citep{synflow} score computes the product of a parameter value and the gradient of the sum of the losses on the minibatch to the parameter:
$$s_\mathtt{synflow}(\theta) = \theta \cdot \frac{\partial \left(\sum_\text{minibatch}\mathcal{L}(\mathbbm{1}; |W|)\right)}{\partial \theta}$$
We also extended \texttt{SYNFLOW} of a single parameter to the saliency of a LoRA module by summation:
\begin{align*}
    S_\mathtt{synflow}(M) &= \sum_{\theta\in M} s_\mathtt{synflow}(\theta) \\
    &= \sum_{\theta\in A} s_\mathtt{synflow}(\theta) + \sum_{\phi\in B} s_\mathtt{synflow}(\phi)
\end{align*}

\subsection{GRAD-NORM}\label{app:proxy:grad-norm} 
A minibatch of training data is fed to the model, and \texttt{GRAD-NORM} computes the Euclidean norm of the loss gradients on a block of parameters:
$$s_\mathtt{gradnorm}(W) = \left\lVert\frac{\partial \mathcal{L}(\mathcal{D}; W)}{\partial W}\right\rVert_2$$
This marks how sensitive the loss is to each block of parameters. We extend this to the saliency of a LoRA module by taking the sum of \texttt{GRAD-NORM} over the matrices:
$$S_\mathtt{gradnorm}(M) = s_\mathtt{gradnorm}(A) + s_\mathtt{gradnorm}(B)$$

\section{LoRA-Adapted Shortcuts}

The detailed definition of LoRA-adapted shortcuts is as follows.

\paragraph{Cross-Layer Shorcut}

This type of shortcut forwards the model's hidden state as:
$$h_{i+1} = s(h_i) + f_i(h_i)$$
where $h_i$ denotes the input hidden state to the $i$th layer of modules, $f_i$ denotes the function of the $i$th layer, and $s$ denotes the current shortcut linear transformation. The function of the layer $f_i$, however, does not necessarily need to be an exact Transformer block as:
$$f_\text{in}(h) = \mathrm{MLP}_i\left(\mathrm{Attention}_i(h)\right)$$
but can also be a ``layer'' recomposed by the sub-modules cutting across a Transformer block boundary, like:
$$f_\text{cut}(h) = \mathrm{Attention}_{i+1}(\mathrm{MLP_i}(h))$$
The two corresponding styles of cross-layer shortcuts, referred to as $s_\text{in}$ and $s_\text{cut}$, are employed in our shortcut-adapted models (as the \textcolor{Green}{green} blocks and the \textcolor{BrickRed}{red} blocks in \Cref{fig: shortcut arch}). 

\paragraph{LayerNorm Inserted After Shortcut}

Layer normalisation needs to be performed after the cross-layer shortcut output is merged into the original hidden state. In this project, the shortcuts are applied to the OPT-350M model, which employs post-layer-normalisation Transformer architecture (layer normalisation is performed after the original residual connection is merged back). For cross-layer shortcuts, given that the original layer normalisation performs an element-wise affine transformation with pre-trained weights, performing another layer normalisation without affine transformation after the original will impact the original layer normalisation's effect, meanwhile, training new weights for a new affine transformation merely on a downstream dataset will be ineffective. Therefore, we re-perform the original layer normalisation after the cross-layer shortcut output is merged back to the hidden state.

Consequently, the original output hidden state $h_{i+1}$ of the $i$th layer:
\begin{align*}
    &a_i = \mathrm{LN}_{1,i}\left( h_i + \mathrm{Attn}_i(h_i) \right) \\
    &h_{i+1} = \mathrm{LN}_{2,i}\left( a_i + \mathrm{FFN}_i(a_i) \right)
\end{align*}
is transformed by the shortcut connections into:
\begin{align*}
    &a_i = \mathrm{LN}_{1,i}\left[ \mathrm{LN}_{1,i}\left( s_{\text{res1},i}(h_i) + \mathrm{Attn}_i(h_i) \right) + s_{\text{cut},i}(a_{i-1}) \right] \\
    &h_{i+1} = \mathrm{LN}_{2,i}\left[ \mathrm{LN}_{2,i}\left( s_{\text{res2},i}(a_i) + \mathrm{FFN}_i(a_i) \right) + s_{\text{in},i}(h_i) \right]
\end{align*}
where $s_{\text{res1},i}, s_{\text{res2},i}, s_{\text{in},i}, s_{\text{cut},i}$ denote the linear projections on the shortcuts, $\mathrm{LN}_{1,i}$ and $\mathrm{LN}_{2,i}$ represent the two layer normalisation layers in the $i$th layer, $\mathrm{Attn}_i$ represents the attention submodule, and $\mathrm{FFN}_i$ represents the feed-forward network submodule.

\section{Additional Experimental Results}
\label{app:additional results}

\begin{figure*}[h]
    \centering
    \begin{subfigure}{\linewidth}
        \includegraphics[width=\linewidth]{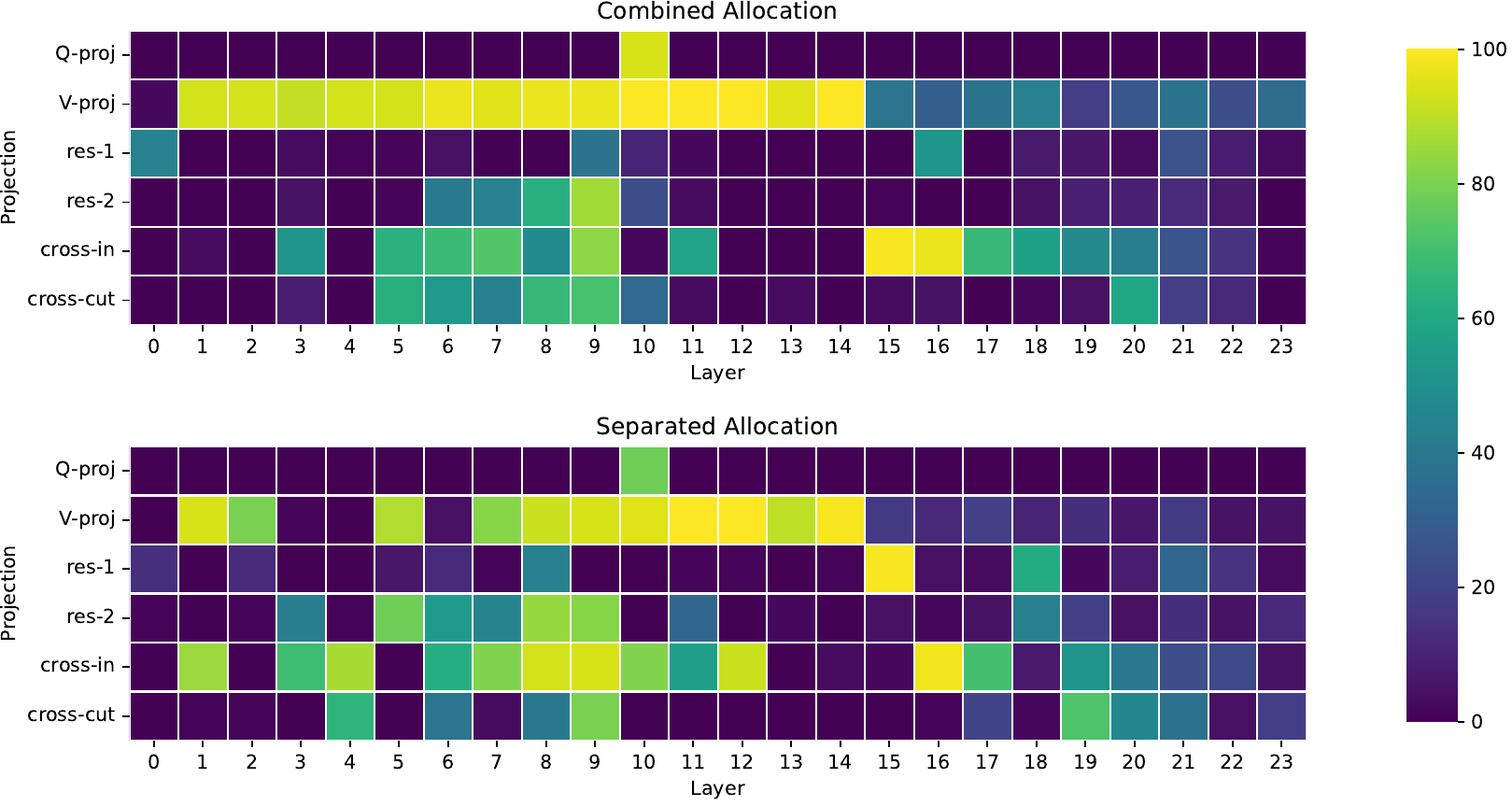}
        \caption{Dynamic HeteroLoRA with modules at $r=8$\\\ }
    \end{subfigure}
    \begin{subfigure}{\linewidth}
        \includegraphics[width=\linewidth]{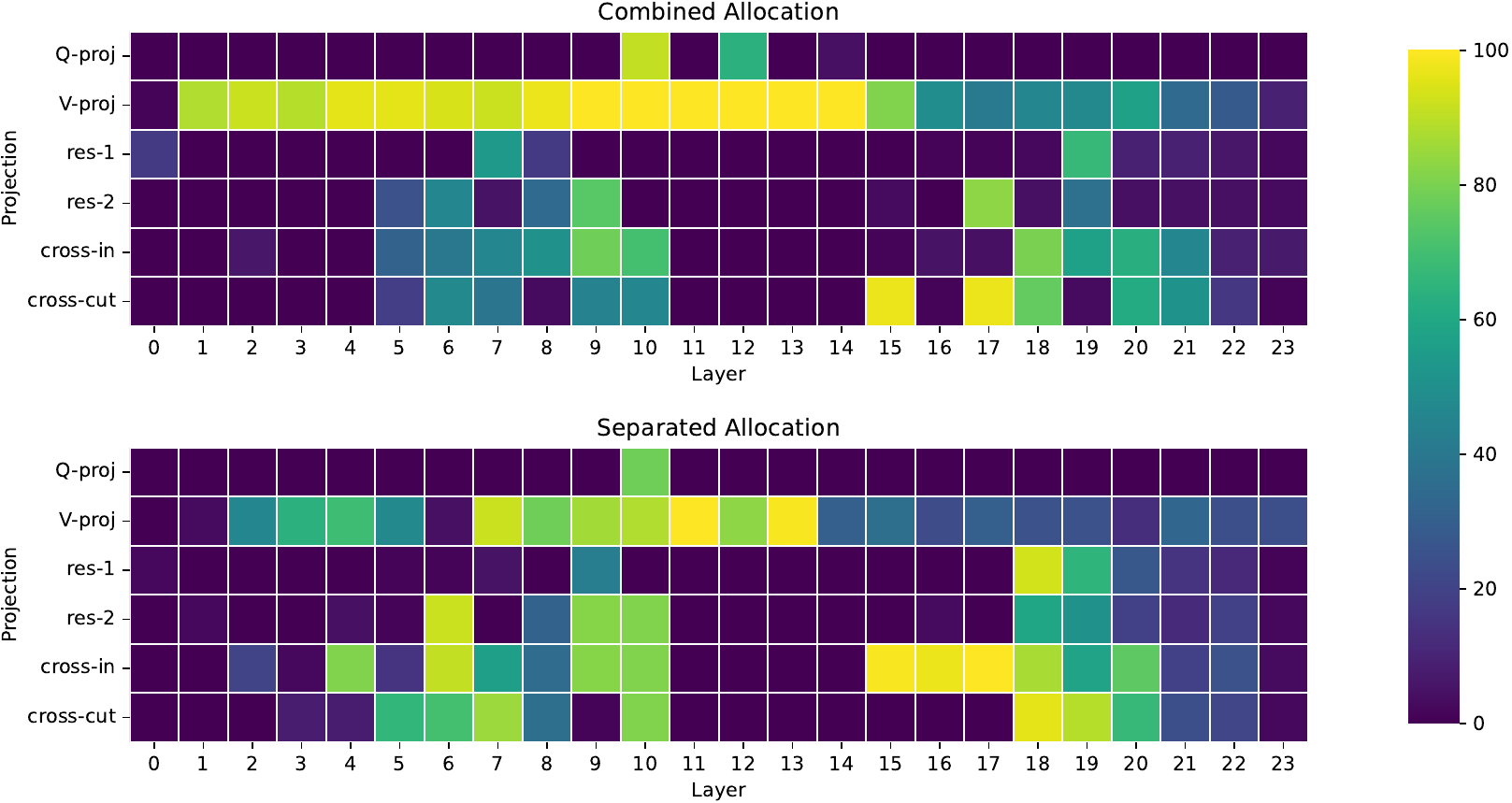}
        \caption{Dynamic HeteroLoRA with modules at $r=32$\\\ }
    \end{subfigure}
    \caption{Frequency of linear projections in every model layer being enabled in the dynamic HeteroLoRA training on RTE with (a) $r=8$ and (b) $r=32$ for LoRA and shortcut modules. For each rank value, combined LoRA allocation is compared to separated LoRA allocation.}
    \label{fig:dyrealloc_heatmap_rte}
\end{figure*}
\begin{figure*}[h]
    \centering
    \begin{subfigure}{\linewidth}
        \includegraphics[width=\linewidth]{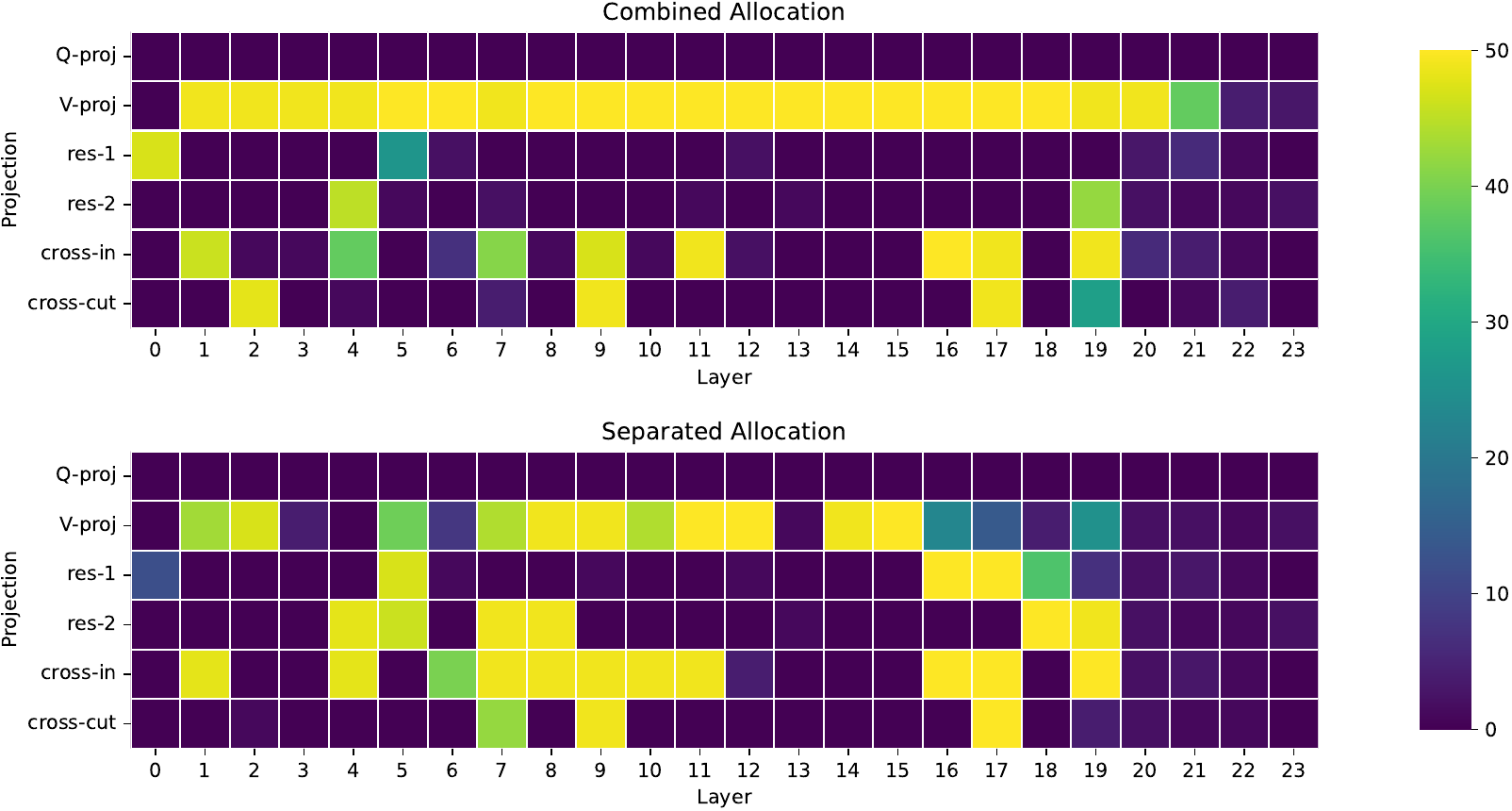}
        \caption{Dynamic HeteroLoRA with modules at $r=8$\\\ }
    \end{subfigure}
    \begin{subfigure}{\linewidth}
        \includegraphics[width=\linewidth]{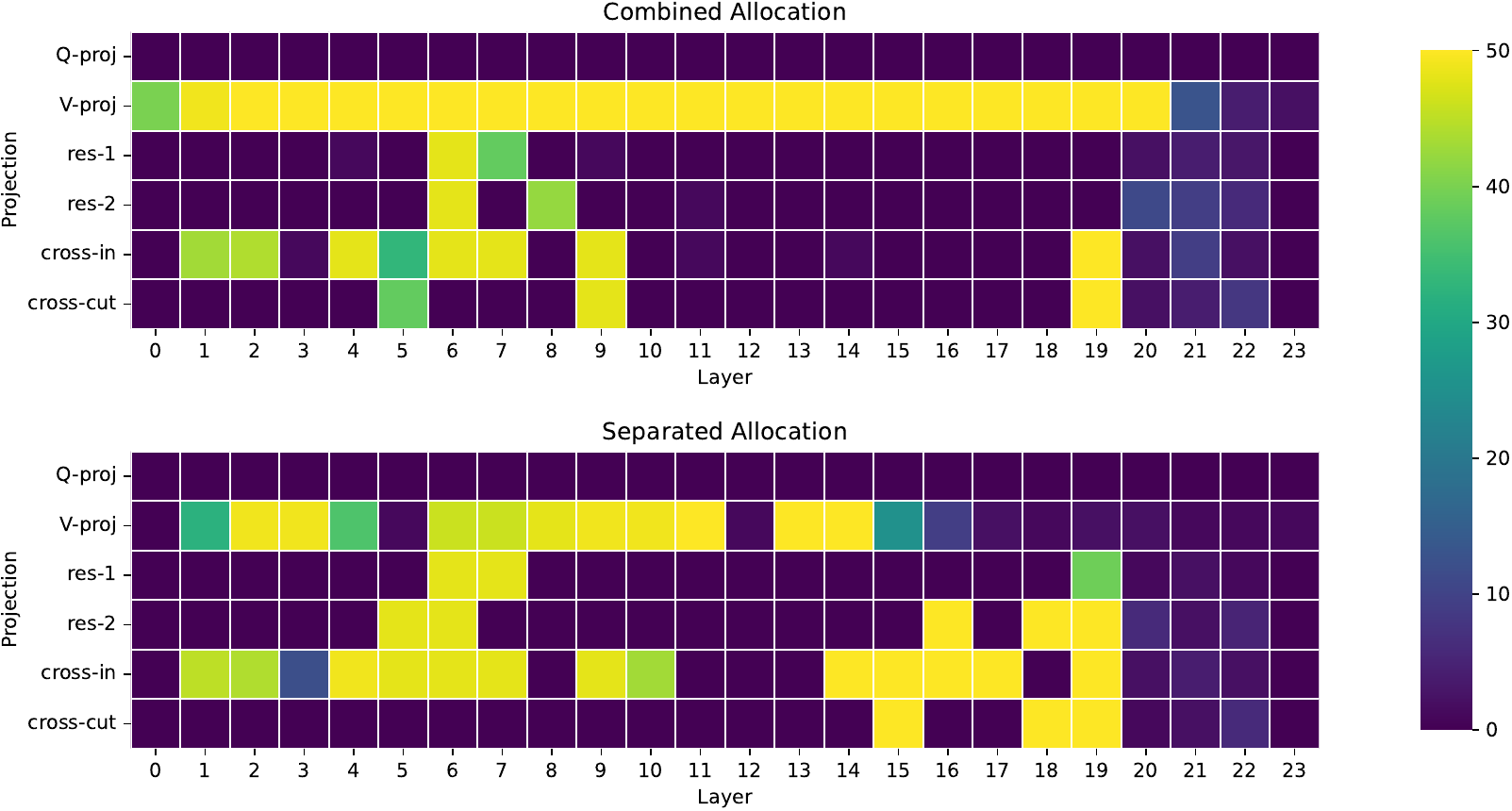}
        \caption{Dynamic HeteroLoRA with modules at $r=32$\\\ }
    \end{subfigure}
    \caption{Frequency of linear projections in every model layer being enabled in the dynamic HeteroLoRA training on SST-2 with (a) $r=8$ and (b) $r=32$ for LoRA and shortcut modules. For each rank value, combined LoRA allocation is compared to separated LoRA allocation.}
    \label{fig:dyrealloc_heatmap_sst2}
\end{figure*}

Dynamic HeteroLoRA is experimented on RTE and SST-2 with the same setup as in \Cref{sec: dyrealloc result}. \Cref{fig:dyrealloc_heatmap_rte} and \Cref{fig:dyrealloc_heatmap_sst2} demonstrate the frequency of each LoRA or shortcut module being enabled over 20 and 10 training epochs on RTE and SST-2, respectively.
Intuitively, frequent LoRA configuration searches give more chances to explore the configuration search space, while a long search interval allows the chosen configuration to be fully trained. \Cref{tab:dyrealloc_N} shows the performance on MRPC and RTE of dynamic HeteroLoRA with various configuration search frequencies. As we can see, no particular performance pattern across the search frequency can be easily observed.

\begin{table*}[h]
    \caption{Dynamic HeteroLoRA with different HeteroLoRA search frequencies per training epoch.}
    \label{tab:dyrealloc_N}
    \centering
    \begin{tabular}{c|>{\centering\arraybackslash}m{0.06\linewidth}>{\centering\arraybackslash}m{0.06\linewidth}>{\centering\arraybackslash}m{0.06\linewidth}>{\centering\arraybackslash}m{0.06\linewidth}|>{\centering\arraybackslash}m{0.06\linewidth}>{\centering\arraybackslash}m{0.06\linewidth}>{\centering\arraybackslash}m{0.06\linewidth}>{\centering\arraybackslash}m{0.06\linewidth}}
    \toprule
    LoRA Ranking Method & \multicolumn{4}{c|}{Combined Allocation} & \multicolumn{4}{c}{Separated Allocation} \\
    Search Freq (per epoch)   & $10$  & $5$  & $2$  & $1$  & $10$   & $5$  & $2$  & $1$  \\
    \midrule
    MRPC (acc)     & 84.6     & 84.3     & 84.1     & 84.3   & 84.6      & 83.7     & 84.6     & 85.1   \\
    RTE (acc)      & 72.3     & 72.9   & 72.3     & 72.7   & 74.5      & 72.8    & 70.1     & 66.9   \\
    \bottomrule
    \end{tabular}
\end{table*}

\end{document}